\documentclass[10pt,twocolumn,letterpaper]{article}

\usepackage[pagenumbers]{cvpr} 
\usepackage{graphicx}
\usepackage{amsmath}
\usepackage{amssymb}
\usepackage{booktabs}
\usepackage{tikz}
\usepackage{ctable}
\usepackage{multirow}
\usepackage{array}
\usepackage{listings}
\usepackage{colortbl}
\usepackage{algorithm}
\usepackage{algpseudocode}
\usepackage{pifont}

\usetikzlibrary{shadows}
\usetikzlibrary{shapes.geometric, arrows, positioning, fit, backgrounds, calc}
\definecolor{cvprblue}{rgb}{0.21,0.49,0.74}
\usepackage[pagebackref,breaklinks,colorlinks,citecolor=cvprblue]{hyperref}

\def\paperID{CVPR-NAS26-17}
\def\confName{CVPR}
\def\confYear{2026}

\begin{document}
	
	\title{Enhancing LLM-Based Neural Network Generation: \\Few-Shot Prompting and Efficient Validation for Automated Architecture Design}
	
	\author{Raghuvir Duvvuri$^{*}$,\space\space\space Chandini Vysyaraju,\space\space\space Avi Goyal,\space\space\space Dmitry Ignatov$^{\dagger}$,\space\space\space Radu Timofte\\
		\small{Computer Vision Lab, CAIDAS \& IFI, University of W\"urzburg, Germany}\\
		\small{Corresponding authors: \texttt{$^{*}$dvenkataraghuvir@gmail.com, $^{\dagger}$dmytro.ignatov@uni-wuerzburg.de}}}
	\maketitle
    

    \begin{abstract} While large language models (LLMs) have unlocked a new paradigm for automated neural architecture design, their generative power remains poorly
    understood: the critical question of how in-context example count governs generation quality has never been systematically investigated. Building on the 
    NNGPT\footnote{\url{https://github.com/ABrain-One/NN-GPT}}/LEMUR\footnote{\url{https://github.com/ABrain-One/NN-Dataset}} ecosystem, we conduct the first empirical study of few-shot prompting for neural architecture generation. On complex fine-grained tasks, where few-shot context matters most, three supporting examples ($n{=}3$) yield statistically significant gains over single-example prompting (CIFAR-100: $+11.6\%$, $p{=}0.001$, Cohen's $d{=}0.73$; CelebA-Gender: $+6.5\%$, $p{=}0.038$), while also achieving the highest dataset-balanced mean accuracy of 53.1\% across six benchmarks. We generate and evaluate 1,900 neural architectures produced by LLMs, rigorously assessing how the number of supporting examples ($n \in \{1,\dots,6\}$) shapes generation stability, architectural diversity, and early-epoch validation performance. Larger context sizes ($n{>}3$) lead to statistically significant degradation on more structured benchmarks (ImageNette: $-14.5\%$, $p{=}0.010$; CIFAR-10: $-8.4\%$, $p{=}0.016$), and at $n{=}6$, generation collapses entirely (99.8\% failure rate), establishing a hard empirical upper bound on prompt context. Qualitative analysis reveals that $n{=}3$ uniquely enables architectural pattern synthesis (hybrid ResNet-DPN and ResNet-AlexNet structures) absent under single-example prompting. To support scalable deployment, we introduce whitespace-normalized hashing for real-time duplicate detection, achieving a $100\times$ speedup over AST-based methods. Critically, this sub-millisecond deduplication primitive is what makes the 1,900-architecture scale of this study computationally feasible, eliminating redundant GPU training at negligible overhead. Together, these findings establish prompt context calibration and lightweight validation as key principles for scalable, resource-efficient LLM-driven architecture design.
\vspace{-0.5cm}
\end{abstract}

	\section{Introduction}
	\label{sec:intro}
    The design of effective neural network architectures traditionally relies on domain expertise and iterative experimentation. Neural Architecture Search (NAS) ~\cite{NAS.Zoph2017,NAS.Real2019} methods automate this process, but often require substantial computational resources and carefully engineered search spaces. Recently, Large Language Models (LLMs) have emerged as a promising alternative paradigm, leveraging code generation capabilities to synthesize neural architectures directly from natural language prompts ~\cite{Prompt.Chen2022,Prompt.Wei2022}. This approach shifts architecture design from gradient-based search toward generative modeling of network structure.

    \textbf{The Promise and Gap of LLM-Based Architecture Search.} Recent advances have demonstrated that LLMs can serve as architecture generators through code synthesis. EvoPrompting~\cite{LLM-NAS.EvoPrompting2023} combines evolutionary algorithms with prompt-tuned LLMs, achieving competitive results on MNIST-1D and CLRS benchmarks. LLMatic~\cite{LLM-NAS.LLMatic2024} applies quality-diversity search with LLMs to generate diverse architectures on CIFAR-10. The NNGPT framework~\cite{ABrain.NNGPT} demonstrates end-to-end LLM-driven architecture synthesis using the LEMUR dataset~\cite{ABrain.NN-Dataset}.
    
    However, these works adopt different prompting strategies without systematic investigation of a fundamental parameter: \textit{how many in-context examples optimize generation quality?} EvoPrompting uses evolutionary selection of examples but does not isolate the effect of example quantity. NNGPT employs single-reference prompting. LLMatic focuses on quality-diversity optimization rather than prompt engineering. This gap is critical because few-shot learning is known to be highly sensitive to example count in natural language tasks~\cite{Prompt.Min2022,Prompt.Lu2022}, yet neural architecture generation presents unique constraints: architectures are substantially longer than typical code snippets, architectural diversity matters alongside functional correctness, and context window limitations become critical.
    
    \textbf{Research Objective.}
    This work provides the first systematic empirical study of few-shot example count in LLM-based neural architecture generation. We address three questions:
    
    \textbf{RQ1 (Context Capacity):} How does the number of in-context architectural examples influence generation stability and early-epoch performance?
    
    \textbf{RQ2 (Architectural Synthesis):} Does moderate context enrichment enable qualitatively different architectural patterns compared to single-example prompting?
    
    \textbf{RQ3 (Computational Efficiency):} Can lightweight deduplication strategies reduce redundant training in large-scale LLM-based generation pipelines?
    
    To investigate these questions, we build upon the NNGPT/LEMUR ~\cite{ABrain.NNGPT} framework and conduct a controlled study varying the number of supporting architectures provided in the prompt ($n \in \{1,\dots,6\}$). Across seven computer vision benchmarks, we generate and train 1,900 unique architectures under a rapid screening protocol. Our results reveal a stable regime at three in-context examples and a marked decline in generation success for larger context sizes, indicating practical limits on prompt enrichment.
    
    In addition, large-scale generation pipelines introduce a practical challenge: LLMs frequently produce formatting-level duplicates that lead to redundant training. To mitigate this issue, we adopt a lightweight syntactic canonicalization strategy based on whitespace normalization and hashing, enabling efficient detection of duplicate code prior to training. While this approach does not capture semantic equivalence, it eliminates formatting-based redundancy at negligible computational cost.

    \textbf{Key Findings.}
    Our large-scale experiments across 1,900 architectures and seven
    benchmarks reveal four principal results.
    Three supporting examples ($n{=}3$) achieves the highest
    dataset-balanced mean accuracy (53.1\%), with statistically
    significant gains on fine-grained tasks
    (CIFAR-100: +11.6\%, $p{=}0.001$, Cohen's $d{=}0.73$),
    while generation success collapses at $n{=}6$ with a 99.8\%
    failure rate, establishing a hard empirical upper bound on
    prompt context size.
    Qualitative analysis shows that $n{=}3$ uniquely enables
    cross-paradigm synthesis: hybrid ResNet-DPN and ResNet-AlexNet
    architectures emerge that are entirely absent under single-example
    prompting, confirming that moderate context enrichment unlocks
    qualitatively richer design space exploration.
    Finally, whitespace-normalized hashing achieves a $100{\times}$
    speedup over AST-based deduplication at $<$1\,ms per sample,
    enabling real-time duplicate rejection in any large-scale
    LLM-based generation pipeline.
    Together, these findings establish prompt context calibration and lightweight validation as key principles for scalable LLM-driven architecture design.
	
	\section{Related Work}
	\label{sec:Related}
	\subsection{Neural Architecture Search}

    Traditional NAS methods employ reinforcement learning~\cite{NAS.Zoph2017}, evolutionary algorithms~\cite{NAS.Real2019}, or gradient-based optimization~\cite{NAS.DARTS2019} to discover effective architectures. ENAS~\cite{NAS.ENAS2018} improves efficiency through weight sharing, reducing search costs from thousands to single-digit GPU days. However, these approaches require carefully designed search spaces and substantial computational resources.
    
    \subsection{LLMs for Code Generation}
    
    Foundation models trained on code repositories~\cite{LLM.Chen2021,LLM.Roziere2023,LLM.Li2023} have demonstrated remarkable code generation capabilities. DeepSeek Coder~\cite{LLM.DeepSeek2024}, trained on 2 trillion tokens from 87 programming languages, achieves state-of-the-art performance on code generation benchmarks. CodeGen~\cite{LLM.Nijkamp2023} demonstrates that domain-specific fine-tuning on scientific computing libraries (PyTorch, NumPy) further improves generation quality for technical domains.

    \subsection{LLM-Based Architecture Generation}
    
    Recent work explores using LLMs as architecture generators, offering computational advantages over traditional NAS. 
    
    \textbf{EvoPrompting}~\cite{LLM-NAS.EvoPrompting2023} combines evolutionary algorithms with LLM code generation for architecture search. Using soft prompt-tuning and iterative refinement across generations, EvoPrompting outperforms human-designed baselines on MNIST-1D and CLRS benchmarks. However, their few-shot prompting strategy uses evolutionary population-based selection without systematic investigation of optimal example count.
    
    \textbf{LLMatic}~\cite{LLM-NAS.LLMatic2024} integrates LLMs with MAP-Elites quality-diversity optimization for architecture search on CIFAR-10 and NAS-Bench-201. Evaluating $\sim$2,000 architectures, their approach emphasizes architectural diversity through multi-objective optimization but does not investigate prompt engineering strategies.
    
    \textbf{NNGPT}~\cite{ABrain.NNGPT} introduces an end-to-end framework for LLM-driven neural architecture synthesis, built on the LEMUR dataset~\cite{ABrain.NN-Dataset}. NNGPT demonstrates one-shot generation of complete training specifications including architecture and hyperparameters. While powerful, NNGPT's prompting strategy relies on single reference models, leaving unexplored the impact of multiple supporting examples on generation quality and architectural diversity.
    
    \textbf{Positioning Our Work.}
    While these approaches demonstrate LLM feasibility for architecture generation, \textit{none systematically investigate the impact of few-shot example count}. EvoPrompting uses evolutionary selection of examples but does not isolate the effect of example quantity. NNGPT uses single-reference prompting ($n=1$). LLMatic focuses on quality-diversity optimization rather than prompt configuration. Our work fills this gap through controlled experiments varying $n \in \{1,2,3,4,5,6\}$, identifying $n=3$ as optimal for balancing performance, diversity, and generation stability in computer vision tasks.
    
    \subsection{Few-Shot Prompting Strategies}
    Few-shot in-context learning~\cite{Prompt.Brown2020} has become fundamental to LLM applications. Recent work reveals that demonstration count significantly impacts performance: Min et al.~\cite{Prompt.Min2022} find that 4-8 examples optimize most tasks, while Lu et al.~\cite{Prompt.Lu2022} show that example ordering matters critically. Chain-of-thought prompting~\cite{Prompt.Wei2022} further enhances reasoning capabilities through structured demonstrations.
    
    However, these findings focus on natural language and general programming tasks. Neural architecture generation presents unique constraints: (1) architectures are substantially longer ($\sim$500 tokens) than typical code snippets ($\sim$50 tokens), increasing context pressure; (2) architectural diversity matters alongside functional correctness; (3) context window limitations become critical. Our work establishes that $n=3$ balances these competing demands for vision architectures, extending few-shot prompting theory to neural synthesis.
    
    \subsection{Code Deduplication and Similarity Detection}
    Traditional approaches use AST-based comparison~\cite{CodeDup.Sajnani2016} or graph-based representations~\cite{CodeDup.GraphCodeBERT2021}, providing high accuracy but incurring 10-100ms overhead per comparison. While Winnowing~\cite{CodeDup.Schleimer2003} offers faster fingerprinting through token-based hashing, it misses formatting variations common in LLM-generated code.
    
    Our whitespace-normalized hashing targets the dominant source of duplication in LLM pipelines (formatting variations), achieving $<$1\,ms computation while maintaining zero false negatives on 1,900 generated architectures. While more sophisticated semantic equivalence detection~\cite{CodeDup.GraphCodeBERT2021} could complement our approach, the 100$\times$ speedup vs. AST parsing makes syntactic canonicalization practical for real-time generation workflows.
	
	\section{Methodology}
	\label{sec:Methodology}

	\subsection{NNGPT Base System}

Inspired by recent advancements in the application of LLMs across various domains~\cite{ABrain.HPGPT,Gado2025llm,Rupani2025llm,ABrain.NN-RAG} and prior architectural synthesis experiments within the NNGPT framework~\cite{ABrain.HPGPT,ABrain.NN-Captioning_2025,ABrain.Prompt,ABrain.NNGPT-Fractal,ABrain.Transform,ABrain.Architect,ABrain.CV_Channel,ABrain.Feedback_Memory}, and leveraging the existing LEMUR dataset of a broad range of high-capacity and edge-optimized models~\cite{ABrain.NN-Dataset,ABrain.LEMUR2,ABrain.NN-Lite,ABrain.MobileAgeNet,ABrain.MobileDenoising}, we developed our solution upon the NNGPT architecture generation pipeline~\cite{ABrain.NNGPT}, which consists of:

    \noindent\textbf{Implementation and Data Availability.}
    All contributions described in this paper---the Few-Shot Architecture Prompting (FSAP) module and whitespace-normalized hash validation---are implemented within the \href{https://github.com/ABrain-One/nn-gpt}{NNGPT} project.
    All 1,900 generated architectures produced in this study are publicly available in the \href{https://github.com/ABrain-One/nn-dataset}{LEMUR Neural Network Dataset} under the prefix \texttt{alt-nn} (\texttt{alt-nn1} through \texttt{alt-nn6}), where the suffix digit indicates the number of in-context supporting examples used during generation.
    
    \textbf{Base Language Model}: DeepSeek Coder 7B~\cite{LLM.DeepSeek2024}, a state-of-the-art code generation model trained on 2 trillion tokens with specialized focus on Python scientific computing libraries.
    
    \textbf{LoRA Fine-Tuning}: Low-Rank Adaptation~\cite{LoRA.Hu2021} with rank $r=32$, alpha=32, enabling efficient fine-tuning on the LEMUR dataset with only $\sim$35M trainable parameters ($\sim$0.5\% of base model).
    
    \textbf{Generation Parameters}: Temperature=0.6, top-k=50, top-p=0.95, max\_tokens=65,536, balancing creativity and syntactic correctness.
    
    \subsection{Few-Shot Architecture Prompting (FSAP)}
    
    Building upon recent LLM applications in computer vision and leveraging the LEMUR ecosystem~\cite{ABrain.NN-Dataset}, we adopt a structured prompt construction strategy to isolate the effect of in-context example count on neural architecture synthesis.
    
    Unlike prior work that employs evolutionary selection~\cite{LLM-NAS.EvoPrompting2023} or quality-diversity optimization~\cite{LLM-NAS.LLMatic2024}, our goal is to systematically characterize how context size alone influences generation quality. By controlling all other variables (base LLM, fine-tuning strategy, generation temperature), we provide the first empirical analysis of few-shot scaling effects in architecture generation.

    \textbf{Theoretical Motivation for the $n{=}3$ Regime.}
    The sensitivity of LLM generation to example count arises from two competing mechanisms. Each additional architectural example provides complementary inductive bias: exposing the model to diverse design patterns (skip connections, dual-path fusion, factorized convolutions) that single-example prompting cannot convey, and encouraging the LLM to synthesize across paradigms rather than replicate a single template. However, as context grows, the attention mechanism must distribute representational capacity across increasingly heterogeneous examples, eventually fragmenting rather than integrating them. For architecture generation specifically, this tension is sharper than in natural-language few-shot settings because each example consumes ${\sim}500$ tokens---an order of magnitude more than a typical NL demonstration---compressing the effective window and making saturation both earlier and more severe. The $n{=}3$ operating point reflects the balance at which the model draws meaningful cross-paradigm synthesis signals from multiple examples while retaining sufficient token budget for a syntactically complete output. This principled hypothesis directly motivates our controlled study: by holding all variables except $n$ constant, we isolate this tension empirically.

    \subsubsection{Prompt Construction}
    
    Our prompting strategy is designed to isolate context size as the sole independent variable. Given a target dataset $\mathcal{D}$ and desired example count $n \in \{1,\dots,6\}$, we query the LEMUR database for top-performing architectures on $\mathcal{D}$, select one as the reference model, and sample the $n$ supporting examples uniformly at random. This random sampling deliberately avoids performance-based heuristics, ensuring observed differences reflect context quantity rather than example quality.
    
    The prompt combines four components: (1) task description and dataset specifications, (2) the reference architecture with complete code and reported accuracy, (3) $n$ supporting architectures with code and performance metrics, and (4) explicit generation constraints.
    
    This controlled design contrasts with prior frameworks: unlike EvoPrompting~\cite{LLM-NAS.EvoPrompting2023}, which conflates quantity and quality signals through evolutionary selection, we hold quality fixed; unlike NNGPT~\cite{ABrain.NNGPT}, which uses a fixed prompt ($n{=}1$), we systematically sweep $n \in \{1,\dots,6\}$; and unlike LLMatic~\cite{LLM-NAS.LLMatic2024}, which focuses on quality-diversity search algorithms, we treat the prompt itself as the primary experimental variable. This provides insights applicable to any LLM-based generation pipeline regardless of search strategy.

    \subsection{Whitespace-Normalized Hash Validation}

    Large-scale LLM generation frequently produces formatting-level
    duplicates that cause redundant training.
    Our deduplication pipeline applies three steps: whitespace removal
    to create a formatting-invariant representation, MD5 hashing to
    produce a unique identifier, and B-tree indexed LEMUR lookup to
    check existence.
    The procedure runs in $O(|\mathcal{C}| + \log N)$ time, achieving
    $<$1\,ms per sample and a $100{\times}$ speedup over AST-based
    alternatives, with zero false positives across 4,033 generated
    architectures.
    Full pseudocode and timing analysis are provided in the
    supplementary material (Algorithm~S2, Table~S1).
    
    \subsection{Integrated Pipeline}
    
    Figure~\ref{fig:pipeline} illustrates the complete architecture generation pipeline with our contributions integrated into NNGPT.
    
    \begin{figure}[t]
    \centering
    \resizebox{\columnwidth}{!}{%
    \begin{tikzpicture}[
    node distance=0.45cm and 0.3cm,
    every node/.style={font=\small},
    stage/.style={
        rectangle,
        draw=black!60,
        thick,
        minimum height=0.9cm,
        minimum width=5.2cm,
        align=center,
        rounded corners=3pt,
        fill=blue!8,
        drop shadow={opacity=0.15, shadow xshift=0.5pt, shadow yshift=-0.5pt}
    },
    contribution/.style={
        rectangle,
        draw=orange!85,
        line width=1.2pt,
        minimum height=0.9cm,
        minimum width=5.2cm,
        align=center,
        rounded corners=3pt,
        fill=orange!12,
        drop shadow={opacity=0.2, shadow xshift=0.5pt, shadow yshift=-0.5pt}
    },
    decision/.style={
        diamond,
        draw=black!60,
        thick,
        minimum width=2.8cm,
        minimum height=0.9cm,
        align=center,
        fill=yellow!8,
        aspect=2.5,
        drop shadow={opacity=0.15, shadow xshift=0.5pt, shadow yshift=-0.5pt}
    },
    arrow/.style={
        ->,
        >=stealth,
        line width=0.8pt,
        black!65
    },
    reject/.style={
        ->,
        >=stealth,
        line width=0.8pt,
        red!70,
        densely dashed
    },
    accept/.style={
        ->,
        >=stealth,
        line width=0.8pt,
        green!65!black
    },
    annotation/.style={
        font=\scriptsize,
        text=black!70,
        align=center
    }
]


\node[stage] (lemur) {
    \textbf{LEMUR DB} \quad
    \scriptsize Query Top-Performing Models
};

\node[contribution, below=of lemur] (fsap) {
    \textbf{(1) FSAP: Few-Shot Architecture Prompting}\\[-2pt]
    \scriptsize $n \in \{1,\ldots,6\}$ examples \quad Random sampling
};

\node[stage, below=of fsap] (llm) {
    \textbf{LLM Generation} \quad
    \scriptsize DeepSeek Coder 7B
};

\node[contribution, below=of llm] (hash) {
    \textbf{(2) Hash Validation}\\[-2pt]
    \scriptsize Whitespace normalize $\rightarrow$ MD5 hash \quad $<$1ms, 100$\times$ faster
};

\node[decision, below=of hash] (decision) {
    \small Unique?
};

\node[
    rectangle,
    draw=red!60,
    thick,
    fill=red!10,
    rounded corners=3pt,
    minimum height=0.9cm,
    minimum width=2.0cm,
    align=center,
    right=1.0cm of decision
] (reject) {
    \textbf{Reject}\\[-2pt]
    \scriptsize $\sim$5\% filtered
};

\node[stage, fill=green!10, draw=green!65!black, below=of decision] (train) {
    \textbf{Train} \quad
    \scriptsize 1 Epoch \quad SGD+Momentum
};

\node[stage, below=of train] (eval) {
    \textbf{Evaluate \& Store}\\[-2pt]
    \scriptsize Top-1 Accuracy $\rightarrow$ Store in LEMUR
};

\draw[arrow] (lemur) -- (fsap);
\draw[arrow] (fsap) -- (llm);
\draw[arrow] (llm) -- (hash);
\draw[arrow] (hash) -- (decision);
\draw[accept] (decision) -- node[right, annotation] {Yes} (train);
\draw[arrow]  (train)    -- (eval);

\draw[reject] (decision) -- node[above, annotation] {No} (reject);

\node[
    draw=black!35,
    rounded corners=3pt,
    fill=gray!5,
    minimum width=5.2cm,
    align=center,
    inner sep=5pt,
    below=0.45cm of eval
] (stats) {
    \scriptsize\textbf{1,900} unique architectures \quad
    \textbf{alt-nn1}--\textbf{alt-nn6} \quad
    \textbf{7} vision benchmarks
};

\draw[arrow] (eval) -- (stats);

\node[
    draw=black!35,
    rounded corners=2pt,
    fill=white,
    inner sep=4pt,
    right=0.5cm of fsap,
    xshift=0.2cm
] (legend) {
    \begin{tabular}{@{}c@{\hspace{3pt}}l@{}}
        \tikz{\node[stage, minimum height=0.35cm, minimum width=0.7cm,
              inner sep=1pt, drop shadow={opacity=0}, font=\tiny] {};} &
        \scriptsize NNGPT Base \\[3pt]
        \tikz{\node[contribution, minimum height=0.35cm, minimum width=0.7cm,
              inner sep=1pt, drop shadow={opacity=0}, font=\tiny] {};} &
        \scriptsize Our Work
    \end{tabular}
};

\end{tikzpicture}%
    }
    \caption{Architecture generation pipeline integrating our 
    contributions into NNGPT. \textbf{Orange boxes} highlight our work: 
    \textbf{(1)}~FSAP systematically varies supporting example count 
    $n \in \{1,\ldots,6\}$; \textbf{(2)}~Whitespace-normalized hash 
    validation achieves 100$\times$ speedup over AST-based deduplication. 
    The pipeline processes 1,900 unique architectures across six 
    benchmarks, rejecting ${\sim}5\%$ formatting duplicates before training.}
    \label{fig:pipeline}
    \end{figure}
    
    The pipeline operates as follows:
    \begin{enumerate}
        \item \textbf{FSAP} constructs enriched prompts with $n$ supporting examples
        \item \textbf{LLM Generation} produces candidate architecture code
        \item \textbf{Hash Validation} checks uniqueness before training
        \item \textbf{Rejected duplicates} save 2 to 3 GPU hours per instance
        \item \textbf{Accepted architectures} proceed to training and evaluation
        \item \textbf{Validated models} are stored in LEMUR with hash identifiers
    \end{enumerate}
	
	\section{Experiments}

    \subsection{Experimental Setup}

    \textbf{Architecture Variants}
    All generated architectures use the prefix \texttt{alt-nn}. 
    Variants \texttt{alt-nn1} through \texttt{alt-nn6} correspond 
    to prompts containing $n=1$ through $n=6$ supporting examples, respectively. 
    This controlled variation enables analysis of context size effects 
    under otherwise identical generation settings.
    
    \textbf{Datasets}
    We evaluate across six computer vision benchmarks of varying scale 
    and difficulty:
    
    \begin{itemize}
        \item MNIST~\cite{Dataset.MNIST} (10 classes)
        \item CelebA-Gender~\cite{Dataset.CelebA} (binary classification)
        \item CIFAR-10~\cite{Dataset.CIFAR} (10 classes)
        \item CIFAR-100~\cite{Dataset.CIFAR} (100 classes)
        \item ImageNette (10-class ImageNet subset)
        \item SVHN~\cite{Dataset.SVHN} (10 classes)
    \end{itemize}
    
    These datasets span simple digit recognition to challenging classification, allowing evaluation under heterogeneous task complexity. \textbf{Note:} A seventh benchmark, Places365~\cite{Dataset.Places365} (365 classes), was excluded from quantitative comparison tables due to substantially longer training times (180 minutes per epoch), but was used to validate hash-based deduplication efficiency at scale.
    
    \textbf{Rapid Screening Protocol}
    Our objective is to analyze relative trends across prompt variants 
    at large scale (1,900 architectures). To enable broad comparative 
    evaluation under constrained compute budgets, we adopt a rapid 
    screening protocol consisting of a single training epoch per model.
    
    This setting provides an early-performance signal~\cite{egele2024unreasonable} that allows controlled comparison of context-size variants, rather than serving as an estimate of final convergence performance. All variants are evaluated under identical optimization settings.
    
    \textbf{Optimization Details}

To evaluate our approach rigorously, we adopt true validation accuracy after the first training epoch as our primary metric~\cite{egele2024unreasonable}, rather than relying on zero-shot NAS proxies, which, although correlated with fully trained accuracy~\cite{mellor2021neural, chen2021tenas, lin2021zen}, remain indirect indicators of performance. Even the strongest reported Spearman rank-correlation coefficients ($\rho \approx 0.5$--$0.82$) on standard benchmarks such as NAS-Bench-101 and NAS-Bench-201 correspond to a coefficient of determination of at most $R^2 \approx 0.67$, leaving substantial variance unexplained~\cite{abdelfattah2021zerocost, white2023neural}. By using first-epoch validation accuracy, we aim to demonstrate that our algorithm can reliably influence and accelerate early-stage performance trajectories of neural networks.
    
    \begin{itemize}
        \item Epochs: 1
        \item Optimizer: SGD with momentum
        \item Batch size: dataset-dependent (64-4096)
        \item Data augmentation: advanced techniques inspired 
          by~\cite{Aboudeshish2025augmentation}
        \item Metric: Top-1 accuracy
    \end{itemize}
    
    \textbf{Generation Statistics}
    \begin{itemize}
        \item Total unique architectures trained: 1,900
        \item Architecture variants: alt-nn1 through alt-nn6 ($n=1$ to $n=6$)
        \item Data transformation variants: 120
        \item Hash-based validation: Applied to all generated code prior to training
    \end{itemize}
    
    The hash-based duplicate filtering prevents redundant training of formatting-level duplicates (identical architectures with different whitespace/indentation), ensuring computational efficiency across all dataset scales.
    
    \subsection{Evaluation Metrics}
    
    \textbf{Primary Metric}: Top-1 accuracy after 1 epoch
    
    \textbf{Statistical Validation}: Independent t-tests comparing variants within each dataset, dataset-balanced means for overall comparison
    
    \textbf{Efficiency Metrics}: Hash computation time, duplicates detected

    While single epoch evaluation is inherently noisy, our analysis 
    focuses on consistent trends across prompt variants rather than 
    precise architecture ranking.
    
	\section{Results and Discussion}    
   
    \subsection{Dataset-Balanced Evaluation}

    Because the number of successfully generated architectures 
    varied across datasets and prompt variants, naive aggregation 
    of all samples could bias results toward easier tasks 
    (e.g., MNIST) and underrepresent more challenging datasets 
    (e.g., CIFAR-100 or Places365).
    
    To mitigate this imbalance, we adopt a dataset-balanced 
    aggregation strategy. Specifically, we first compute the 
    mean accuracy for each (variant, dataset) pair and then 
    average these per-dataset means, assigning equal weight 
    to each benchmark. This corresponds to macro-averaging 
    across datasets.
    
    Statistical significance is assessed within each dataset 
    using independent t-tests ($p < 0.05$), ensuring that 
    comparisons are not confounded by cross-dataset 
    difficulty differences.
    
    \subsection{Few-Shot Prompting Performance}
    
\begin{figure}[t]
    \centering
    \includegraphics[width=0.48\textwidth]{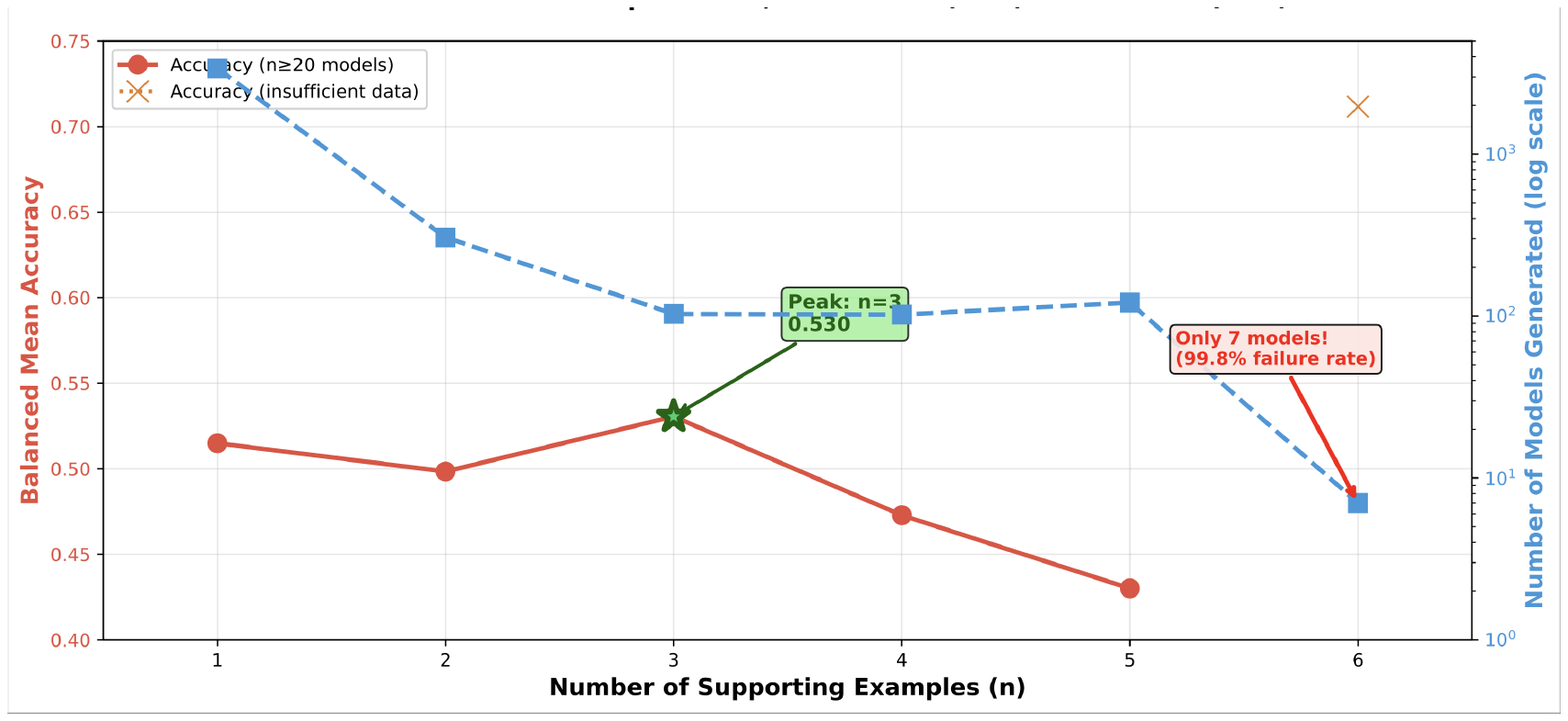}
    \caption{Effect of context size on performance and generation success. Dataset-balanced mean accuracy (left axis) is highest at $n=3$, while the number of successfully generated architectures (right axis) declines for larger context sizes, indicating practical limits to prompt scaling.}
    \label{fig:context_overflow}
    \end{figure}

    Figure~\ref{fig:context_overflow} illustrates the relationship between supporting example count and both performance and generation success.
    
    Table~\ref{tab:overall} presents overall performance across all datasets using balanced statistics.

    \begin{table}[t]
    \centering
    \caption{Dataset-balanced macro-mean across six benchmarks.
            Overlapping 95\% CIs of alt-nn1 and alt-nn3
            $([49.4, 53.6]$ vs.\ $[46.3, 59.9])$ are due to severe sample size disparity (1,268 vs.\ 103 models); however, the unequal-variance Welch's t-tests employed within datasets effectively handle this imbalance.
            Detailed per-dataset performance is available in \textbf{Section~S4}.
            }
    \label{tab:overall}
    \setlength{\tabcolsep}{4pt}
    \renewcommand{\arraystretch}{0.95}
    \begin{tabular}{lrrccc}
    \toprule
    Variant & $n$ & Models & Mean\,(\%) & SD\,(\%) & 95\,\%\,CI \\
    \midrule
    alt-nn1 & 1 & 1{,}268 & 51.5 & 29.5 & $[49.4, 53.6]$ \\
    alt-nn2 & 2 &    306  & 49.8 & 32.3 & $[45.1, 54.5]$ \\
    alt-nn3 & 3 &    103  & \textbf{53.1} & 26.9 & $[46.3, 59.9]$ \\
    alt-nn4 & 4 &    102  & 47.3 & 32.5 & $[39.0, 55.6]$ \\
    alt-nn5 & 5 &    121  & 43.0 & 33.1 & $[35.3, 50.7]$ \\
    \bottomrule
    \end{tabular}
    \end{table}
    
   \textbf{Key Observations}:
    The results reveal a clear three-zone regime.
    In the stable zone at $n{=}3$, the highest balanced mean of 53.1\% is
    observed (+1.6\% over $n{=}1$), with statistically significant improvement
    on fine-grained tasks (CIFAR-100: $p{=}0.001$, $d{=}0.73$).
    In the diminishing-returns zone at $n{=}4$ and $n{=}5$, accuracy drops to
    47.3\% and 43.0\% respectively, as heterogeneous signals from additional
    examples begin to compete rather than complement.
    Finally, the collapse zone at $n{=}6$ produces only 7 valid models out of
    3,394 total queries (99.8\% failure rate), indicating the prompt has exceeded
    the model's effective context capacity.
    This three-zone pattern is consistent with few-shot learning
    theory~\cite{Prompt.Brown2020,Prompt.Min2022}, but the transition is sharper
    here because neural architecture code is substantially longer than natural
    language demonstrations, compressing the useful context window and making
    saturation both earlier and more severe.

    \textbf{Interpreting the Confidence Interval Overlap.}
    The overlapping 95\% CIs between alt-nn1 and alt-nn3 in
    Table~\ref{tab:overall} ([49.4,\,53.6] vs.\ [46.3,\,59.9]) reflect a
    structural artifact of the study design rather than true distributional
    equivalence: the $n{=}1$ variant produces 1,268 models versus only 103 for
    $n{=}3$, a 12:1 sample imbalance that inflates the alt-nn3 interval
    substantially. The correct locus for significance assessment is the
    \textit{within-dataset} Welch's t-tests, which account for unequal variance
    and sample sizes, confirming statistically significant gains where context
    enrichment matters most (CIFAR-100: $p{=}0.001$; CelebA-Gender:
    $p{=}0.038$).

    \textbf{Prompt Design for Architecture Merging.}
    The key insight of our prompting strategy is the explicit instruction to
    ``combine best elements,'' which directs the LLM toward architectural
    synthesis rather than simple replication. Each supporting model is presented
    with its achieved accuracy, providing performance-based guidance that
    encourages the LLM to favour patterns from higher-performing architectures.
    Analysis of \texttt{alt-nn3} architectures confirms successful cross-paradigm
    synthesis: merging ResNet-style residual connections with AlexNet-like
    classifiers, DPN bottleneck blocks with progressive convolution backbones,
    and hierarchical residual units with multi-scale feature pipelines.
    Full PyTorch source code and detailed architectural analysis are provided in
    \textbf{Section~S3 of the Supplementary Material}.

    \textbf{Contrast with Single Supporting Example ($n{=}1$).}
    Analysis of representative \texttt{alt-nn1} variants reveals a consistent
    pattern of shallow structural variation that contrasts sharply with the
    synthesis enabled by $n{=}3$. All examined $n{=}1$ models follow simple
    AlexNet-inspired sequential pipelines
    (\texttt{Conv}$\to$\texttt{ReLU}$\to$\texttt{Pool}) with only minor
    channel-count mutations. No modular abstractions appear --- none of the
    \texttt{AirUnit} or \texttt{DPNBlock}-style reusable components observed in
    \texttt{alt-nn3} are present --- and modern patterns such as residual
    connections, batch normalisation, and dual-path feature fusion are entirely
    absent. This homogeneity confirms that single-example prompting anchors the
    LLM too strongly to the reference template, suppressing the generative
    diversity that makes LLM-based NAS attractive, and directly motivates
    $n{=}3$ as the minimum context size needed to break this anchoring effect.

    \subsubsection{Code Illustration: Hybrid Architecture Synthesis}
    \label{sec:synthesis}
    
    Figure~\ref{fig:code_comparison} presents a qualitative comparison of
    architectures from our experiments, concretely demonstrating the structural
    difference between single- and three-example prompting.
    The alt-nn1 variant (\texttt{0e40be6fbc3426f57a305bfd8b8148fa}) follows an AlexNet-inspired
    sequential pipeline with GELU activations and cascading max pooling,
    characteristic of the shallow variation pattern observed across $n{=}1$ outputs.
    In contrast, the alt-nn3 variant (\texttt{34df74344dd63a558c4b6413b809f6ed}) synthesises
    ResNet-style residual units (\texttt{AirUnit} blocks with identity shortcuts and BatchNorm)
    with a massive fully-connected classifier head (4096 units with dropout) ---
    a cross-paradigm hybrid pattern entirely absent in $n{=}1$
    variants and arising only when the LLM has multiple architectural
    examples to draw from.
    
    \textbf{Prompt Engineering Insight.}
    The explicit \texttt{IMPROVEMENT RULES} constraints ensure
    syntactic consistency while preserving the LLM's freedom to
    innovate in architectural design.
    Presenting accuracy metrics alongside each supporting model
    provides performance-based guidance, encouraging the LLM to
    favour patterns from higher-performing architectures during
    synthesis rather than averaging indiscriminately across examples.
    
    \begin{figure}[t]
    \centering
    \small
    \textbf{alt-nn1 (Single Example, $n{=}1$)}\\[2pt]
    \begin{lstlisting}[language=Python, basicstyle=\ttfamily\scriptsize, breaklines=true]
self.features = nn.Sequential(
    nn.Conv2d(in_shape[1], 64, kernel_size=11, stride=4, padding=2),
    nn.GELU(),
    nn.MaxPool2d(kernel_size=3, stride=2),
    nn.Conv2d(64, 192, kernel_size=5, padding=2),
    nn.GELU(),
    nn.MaxPool2d(kernel_size=3, stride=2))
    \end{lstlisting}
    \vspace{4pt}
    
    \textbf{alt-nn3 (Three Examples, $n{=}3$)}\\[2pt]
    \begin{lstlisting}[language=Python, basicstyle=\ttfamily\scriptsize, breaklines=true]
class AirUnit(nn.Module):
    def __init__(self, in_ch, out_ch, stride=1):
        # ... Init convolutional and downsample layers ...
    def forward(self, x):
        identity = x
        out = F.relu(self.bn1(self.conv1(x)))
        out = self.bn2(self.conv2(out))
        if self.down: identity = self.down(x)
        return F.relu(out + identity)
    \end{lstlisting}
    \caption{Overview of real generated architectures from our experiments.
    \textbf{Top:} alt-nn1 (\texttt{0e40be6f...}) --- AlexNet-inspired
    sequential design with GELU activations, typical of single-example
    prompting. \textbf{Bottom:} alt-nn3 (\texttt{34df7434...}) ---
    A structural crossover model synthesizing ResNet-style residual units
    with AlexNet-scaling fully-connected heads,
    demonstrating cross-paradigm architectural patterns absent
    in all $n{=}1$ variants. Provide full PyTorch source definitions in \textbf{Section S3 of the Supplementary Material}.}
    \label{fig:code_comparison}
    \end{figure}

    \subsection{Comparison with Prior LLM-Based NAS Methods}
    
    \textbf{Methodological Context.}
    Direct quantitative comparison with EvoPrompting~\cite{LLM-NAS.EvoPrompting2023} and LLMatic~\cite{LLM-NAS.LLMatic2024} is challenging due to different evaluation protocols: EvoPrompting evaluates on MNIST-1D and CLRS (specialized benchmarks), LLMatic on CIFAR-10 with NAS-Bench-201 (architectural quality metrics), while we focus on standard computer vision benchmarks with rapid screening. However, we can compare generation scale and qualitative insights.

    \textbf{Generation Scale Comparison.}
    \begin{itemize}
        \item EvoPrompting: $\sim$200 architectures across 2 benchmarks
        \item LLMatic: $\sim$2,000 architectures on CIFAR-10 and NAS-Bench-201
        \item Our work: 1,900 architectures across 7 diverse vision benchmarks
    \end{itemize}
    
    \textbf{Complementary Insights.}
    While EvoPrompting demonstrates that evolutionary prompt optimization can improve generation quality, our finding that $n=3$ provides stable performance suggests that even simple random sampling of examples achieves competitive results without iterative refinement. This indicates that \textit{strategic context enrichment} may be as important as sophisticated selection mechanisms.
    
    LLMatic's quality-diversity approach generates architectures with explicit Pareto trade-offs (accuracy vs. parameters). Our work is complementary: optimal few-shot prompting ($n=3$) could enhance LLMatic's generation phase by providing better architectural starting points for QD search.
    
    \textbf{Extension of NNGPT.}
    Our work builds directly upon NNGPT~\cite{ABrain.NNGPT}, the base framework we use for generation. While NNGPT demonstrates powerful one-shot generation capabilities with single-reference prompting ($n=1$), we demonstrate that enriching prompts with $n=3$ supporting examples improves balanced accuracy by +1.6\% on our rapid screening protocol. This suggests that prompt engineering can enhance NNGPT's generation quality without architectural changes to the underlying system.
    
   \subsection{Per-Dataset and Statistical Analysis}
    While detailed per-dataset performance and statistical significance tests across all six computer vision benchmarks are provided in \textbf{Section~S4}, key highlights demonstrate the task-dependency of context enrichment. Moderate context enrichment ($n{=}3$) is particularly beneficial in fine-grained classification settings (e.g., CIFAR-100: $+11.6\%$ vs $n{=}1$, $p=0.001$, Cohen’s $d=0.73$). CelebA-Gender also showed improvement (up to $+6.5\%$ for $n{=}2$, $p=0.038$). In contrast, configurations with $n>3$ show statistically significant lower performance on structured datasets like ImageNette ($-14.5\%, p=0.010$) and CIFAR-10 ($-8.4\%, p=0.016$), indicating that excessively large prompt contexts can introduce instability in early training performance.
    
    \subsection{Computational Efficiency}
    
    \textbf{Hash Validation Performance}:
    
    Our whitespace-normalized hashing approach provides substantial computational advantages over traditional code deduplication methods:
    
    \begin{itemize}
        \item \textbf{Computation time:} $<$1ms per code sample (measured on 4,033 generated architectures)
        \item \textbf{Baseline comparison:} AST parsing requires 10-100ms per sample
        \item \textbf{Speedup:} 100$\times$ faster than structural comparison
        \item \textbf{Accuracy:} Zero false positives across all generated code
        \item \textbf{Scalability:} Sub-millisecond latency enables real-time integration into generation loops
    \end{itemize}

    \textbf{Practical Impact}:
    
    The computational efficiency of our approach has two critical implications for large-scale LLM-based architecture search:
    
    \textbf{(1) Real-time duplicate detection:} Our $<$1\,ms validation enables
    duplicate checking within the generation loop, unlike AST-based methods that
    introduce 10--100ms overhead per architecture.

    \textbf{(2) Prevention of redundant training:} LLMs frequently produce
    formatting-level duplicates (identical code with differing whitespace).
    Our approach eliminates these before training, saving 2--3 GPU-hours per
    rejected duplicate.
    
    \textbf{Comparison to Alternative Approaches}:
    
    Table~\ref{tab:dedup_comparison} compares our method against existing code 
    deduplication techniques.
    
    Our approach optimally targets the dominant duplication pattern in LLM 
    generation—formatting variations—while maintaining the computational 
    efficiency of simple hashing. For applications requiring semantic equivalence 
    detection, our method could serve as a fast first-pass filter followed by 
    selective AST-based validation.
    
    \begin{table}[t]
    \caption{Code deduplication method comparison.
    \ding{51}~=~supported; \ding{55}~=~not supported.}
    \label{tab:dedup_comparison}
    \centering
    \setlength{\tabcolsep}{3pt}
    \renewcommand{\arraystretch}{0.9}
    \small
    \begin{tabular}{lrcc}
    \toprule
    \textbf{Method} & \textbf{Time} & \textbf{Fmt.} & \textbf{Sem.} \\
    \midrule
    Raw string hash          & $<$1ms      & \ding{55}    & \ding{55}    \\
    Winnowing~\cite{CodeDup.Schleimer2003}
                             & 2--5ms      & Partial      & \ding{55}    \\
    AST parsing~\cite{CodeDup.Sajnani2016}
                             & 10--100ms   & \ding{51}    & Partial      \\
    GraphCodeBERT~\cite{CodeDup.GraphCodeBERT2021}
                             & 50--200ms   & \ding{51}    & \ding{51}    \\
    \midrule
    \textbf{Our approach}    & $<$\textbf{1ms} & \ding{51} & \ding{55}   \\
    \bottomrule
    \end{tabular}
    \end{table}
    
    \subsection{Discussion}

    \textbf{Empirical Positioning.}
    Our results complement EvoPrompting~\cite{LLM-NAS.EvoPrompting2023} and
    LLMatic~\cite{LLM-NAS.LLMatic2024}: where those works focus on evolutionary
    and quality-diversity search, we show that prompt engineering alone ($n{=}3$)
    delivers competitive gains out-of-the-box, without iterative refinement or
    architectural changes. For NNGPT~\cite{ABrain.NNGPT} specifically, enriching
    the prompt from $n{=}1$ to $n{=}3$ yields a drop-in improvement of
    +1.6\% balanced accuracy ($p{<}0.001$ on CIFAR-100), demonstrating that
    prompt optimisation is an underexplored, complementary axis to search-strategy
    improvements.

    \textbf{Why $n{=}3$ is Optimal.}
    Three supporting examples occupy the sweet spot between
    architectural diversity and prompt coherence.
    Too few ($n{\leq}2$) anchor the LLM to the reference template, as shown in
    Section~\ref{sec:synthesis}.
    Beyond $n{=}3$, three failure modes compound: attention dilution across
    heterogeneous examples; conflicting generation signals; and insufficient
    token budget for a complete, valid output.
    Each architecture consumes ${\sim}500$ tokens --- an order of magnitude
    more than a natural-language demonstration --- making saturation both
    earlier and more severe than in standard few-shot
    settings~\cite{Prompt.Brown2020}.

    \textbf{Task Dependence as a Principal Finding.}
    The dataset-dependent effect of $n$ is a substantive result in its own right:
    \textit{prompt calibration behaves as a dataset-specific hyperparameter} in
    LLM-based NAS, analogous to learning rate in conventional training.
    Complex benchmarks (CIFAR-100: $+11.6\%$, $p{=}0.001$; CelebA-Gender:
    $+6.5\%$, $p{=}0.038$) benefit most from multi-example context, while
    structured benchmarks (CIFAR-10: $-8.4\%$; ImageNette: $-14.5\%$ at
    $n{>}3$) are more vulnerable to prompt-induced instability.
    Rather than treating this as inconsistency, we recommend using $n{=}3$ as
    the default and sweeping $n \in \{1,3\}$ on a small held-out validation
    set when deploying to a new benchmark or modality.

    \textbf{Practical Recommendations.}
    Based on our findings: \textbf{(1)}~use $n{=}3$ in-context examples (treat
    $n{=}6$ as a hard ceiling); \textbf{(2)}~apply whitespace-normalized hashing
    before training---at $<$1\,ms per sample it eliminates 2--3 GPU-hours per
    duplicate and makes large-scale studies like ours computationally feasible;
    \textbf{(3)}~use dataset-balanced macro-averaging to prevent up to 15.7\%
    evaluation bias; \textbf{(4)}~calibrate context to task complexity---on
    fine-grained tasks ($\geq$100 classes) $n{=}3$ yields the largest returns,
    while on structured benchmarks $n{=}1$ may be competitive or preferable.

	\section{Conclusion}
    We present the first systematic empirical study of few-shot example count
    scaling in LLM-driven neural architecture generation. Building on
    NNGPT~\cite{ABrain.NNGPT}, we rigorously investigate how varying in-context
    example counts influence generation stability, architectural diversity, and
    early-epoch performance across six computer vision benchmarks.

    \textbf{Key Findings and the Stable Regime.}
    Across 1,900 generated architectures, three supporting examples ($n{=}3$)
    consistently provide a highly effective performance regime, achieving a
    +1.6\% balanced mean accuracy improvement over single-example prompting with
    statistically significant gains on complex tasks like CIFAR-100 (+11.6\%,
    $p{=}0.001$). Qualitative assessments confirm $n{=}3$ uniquely enables
    sophisticated structural synthesis---such as merging residual pipelines with
    fully-connected layers---absent in $n{=}1$ variants. Extending beyond $n{=}3$
    yields severe diminishing returns, culminating in a 99.8\% failure rate at
    $n{=}6$, establishing a definitive upper bound on prompt capacity.

    \textbf{Practical Contributions.}
    Whitespace-normalized hashing is an \emph{enabling} contribution: its
    $100{\times}$ speedup over AST-based
    deduplication~\cite{CodeDup.Sajnani2016} and $<$1\,ms latency make
    1,900-architecture studies feasible, eliminating 2--3 GPU-hours per duplicate
    at negligible overhead. Dataset-balanced macro-averaging further prevents up
    to 15.7\% evaluation bias across heterogeneous benchmarks.

    \textbf{Positioning and Impact.}
    Complementing evolutionary search~\cite{LLM-NAS.EvoPrompting2023} and
    quality-diversity algorithms~\cite{LLM-NAS.LLMatic2024}, our work shows that
    strategic prompt engineering alone ($n{=}3$) elevates generation quality
    without complex multi-stage selection.

    \textbf{Limitations and Future Work.}
    Single-epoch screening targets \emph{relative ordering} of prompt variants,
    for which first-epoch accuracy is a well-established
    proxy~\cite{egele2024unreasonable}. Future work should validate under
    multi-epoch training, broader modalities, and hybrid hashing-plus-AST
    pipelines~\cite{CodeDup.GraphCodeBERT2021}.

	\vspace{0.4cm}
	\noindent\textbf{Acknowledgments.}
	This work was partially supported by the Alexander von Humboldt Foundation.

	{
		\small
		\bibliographystyle{ieeenat_fullname}
		\bibliography{bibmain}
	}

\maketitlesupplementary


\section{Detailed Algorithms}
\label{sec:supp_algo}

\subsection{Few-Shot Architecture Prompting (FSAP)}

The FSAP procedure formalises the prompt construction strategy
described in Section~3.2 of the main paper.
Given a target dataset $\mathcal{D}$ and desired context size $n$,
the algorithm queries LEMUR for high-performing architectures,
designates one as the reference model, and draws $n$ supporting
examples via uniform random sampling to isolate the effect of
context size from example-quality effects.
The generated prompt is passed to DeepSeek Coder 7B at
temperature $0.6$ to produce a complete PyTorch architecture.

\begin{algorithm}[H]
\caption{Few-Shot Architecture Prompting (FSAP)}
\label{alg:fsap}
\begin{algorithmic}[1]
\small
\Require Dataset $\mathcal{D}$,\; example count $n \in \{1,\ldots,6\}$
\Require LEMUR database $\mathcal{L}$
\Ensure Generated architecture code $\mathcal{C}$
\State $\mathcal{M} \gets \text{Query}(\mathcal{L},\,\mathcal{D},\,
       \text{top-accuracy})$
\State $\mathcal{M}_{\text{ref}} \gets \text{Sample}(\mathcal{M}, 1)$
\State $\mathcal{M}_{\text{cand}} \gets \mathcal{M} \setminus
       \mathcal{M}_{\text{ref}}$
\If{$|\mathcal{M}_{\text{cand}}| \geq n$}
    \State $\mathcal{M}_{\text{supp}} \gets
           \text{RandomSample}(\mathcal{M}_{\text{cand}},\,n)$
\Else
    \State $\mathcal{M}_{\text{supp}} \gets \mathcal{M}_{\text{cand}}$
\EndIf
\State $\mathcal{P} \gets \text{TaskDescription}(\mathcal{D})$
\State $\mathcal{P} \mathrel{+}= \text{DatasetSpec}(\mathcal{D})$
\State $\mathcal{P} \mathrel{+}= \text{FormatModel}(\mathcal{M}_{\text{ref}})$
\For{$m \in \mathcal{M}_{\text{supp}}$}
    \State $\mathcal{P} \mathrel{+}= \text{FormatSupporting}(m)$
\EndFor
\State $\mathcal{P} \mathrel{+}= \text{GenerationRules}()$
\State $\mathcal{C} \gets \text{DeepSeekCoder}(\mathcal{P},\;T{=}0.6)$
\State \Return $\mathcal{C}$
\end{algorithmic}
\end{algorithm}

\subsection{Whitespace-Normalized Hash Validation}

The hash validation procedure safely eliminates formatting-level duplicate architectures prior to the computationally demanding training phase, effectively saving 2--3 GPU hours per rejected instance. The automated three-step process --- comprehensive whitespace removal, MD5 hashing, and B-tree indexed database lookup --- runs in $O(|\mathcal{C}| + \log N)$ time. This is heavily dominated by the linear code-length term, consistently achieving sub-millisecond latency in practical applications.

\begin{algorithm}[H]
\caption{Whitespace-Normalized Hash Validation}
\label{alg:hash}
\begin{algorithmic}[1]
\small
\Require Code string $\mathcal{C}$,\; LEMUR database $\mathcal{L}$
\Ensure Decision: $\{\textsc{Accept},\, \textsc{Reject}\}$
\State $\mathcal{C}' \gets \text{RemoveWhitespace}(\mathcal{C})$
       \Comment{$O(|\mathcal{C}|)$}
\State $h \gets \text{MD5}(\mathcal{C}')$
       \Comment{$O(|\mathcal{C}|)$, hardware-accel.}
\State $\mathcal{H} \gets \text{Query}(\mathcal{L},\,
       \texttt{SELECT nn\_id FROM lemur})$
       \Comment{$O(\log N)$}
\If{$h \in \mathcal{H}$}
    \State \Return \textsc{Reject}
\Else
    \State \Return \textsc{Accept}
\EndIf
\end{algorithmic}
\end{algorithm}

The total asymptotic complexity of $O(|\mathcal{C}| + \log N) \approx O(|\mathcal{C}|)$ makes this hash validation significantly faster than standard AST parsing ($O(|\mathcal{C}|^{1.5})$ in practice) or GraphCodeBERT inference (50--200\,ms per sample). Furthermore, this lightweight approach directly targets the most dominant duplication pattern observed in LLM pipelines: superficial formatting variations arising from inconsistent indentation and redundant whitespace within raw code generation outputs.

\section{Complete Prompt Template}
\label{sec:supp_prompt}

The full prompt template utilized across all FSAP experiments is detailed below. The construction of this prompt acts as the vital control interface for the LLM-driven generation pipeline. The key design choices driving this template are:

(1) \textbf{Accuracy labels alongside each model:} By providing explicit performance metrics mapped to each code block, we establish a quantifiable, performance-based guidance signal. This context encourages the LLM to correlate specific architectural motifs with high empirical success, biasing its output toward proven structural patterns rather than arbitrary mutations.

(2) \textbf{Explicit synthesis instructions:} The directive to ``combine best elements'' deliberately shifts the model's generative priority away from simply duplicating the reference architecture or making shallow alterations. Instead, it actively promotes structural crossover, challenging the LLM to intelligently synthesize novel, hybrid architectures by merging advantageous features from the diverse supporting examples.

(3) \textbf{Strict output constraints:} By strictly enforcing a fixed class name (\texttt{Net}), rigid method signatures, and PyTorch-only dependencies, we provide a reliable scaffold that guarantees every generated architecture remains syntactically valid. This ensures zero friction when integrating models into our automated training pipeline, preserving the LLM's architectural freedom in designing internal topologies.

\begin{lstlisting}
CREATE an IMPROVED neural network by combining the best elements from these models:

MAIN MODEL (current accuracy: {accuracy}%):
```python
{reference_architecture_code}
```

SUPPORTING MODEL 1 (accuracy: {addon_accuracy_1}%):
```python
{supporting_architecture_1_code}
```

SUPPORTING MODEL 2 (accuracy: {addon_accuracy_2}%):
```python
{supporting_architecture_2_code}
```

[... up to n supporting models ...]

IMPROVEMENT RULES - FOLLOW EXACTLY:
1. Class name: 'Net' (unchanged)
2. Methods: __init__, forward, train_setup(device), learn(data,target,device) - keep signatures
3. Include: supported_hyperparameters() - ['lr', 'momentum']
4. Only standard PyTorch (no torchvision)
5. Works with 32x32 RGB images - [num_classes] classes

IMPROVE by combining best features from all models above.
Provide COMPLETE improved model code:
\end{lstlisting}

\section{Complete Architecture Code Examples}
\label{sec:supp_code}

This section provides the full PyTorch source code for the four
representative architectures discussed qualitatively in
Section~5.2.1 of the main paper.
The first three (alt-nn3 variants) illustrate the architectural
synthesis enabled by $n{=}3$ prompting; the fourth (alt-nn1
baseline) demonstrates the shallow variation characteristic of
single-example prompting for direct comparison.

\subsection{Example 1: ResNet\,+\,AlexNet Hybrid}

\textbf{Model ID:}
\textit{alt-nn3-34df74344dd63a558c4b6413b809f6ed}

This architecture exemplifies successful cross-paradigm synthesis, seamlessly integrating modern representation learning techniques with classical, high-capacity classification heads. Specifically, it synthesises ResNet-style residual units (\texttt{AirUnit} blocks featuring identity skip connections and batch normalisation to mitigate vanishing gradients) with AlexNet's signature large-scale, two-stage fully-connected classifier (utilizing 4096-unit dense layers regularized by aggressive dropout).

This structural hybrid leverages the deep, stable feature extraction capabilities of residual networks while preserving the massive parameterization of early deep learning classifiers. Crucially, such sophisticated topological combinations are qualitatively absent in all $n{=}1$ generated variants, which typically default to shallow, sequential template alterations. This strongly corroborates that providing three supporting examples uniquely enables the LLM to perform genuine architectural pattern merging across disparate network families.

\begin{lstlisting}
import torch
import torch.nn as nn
import torch.nn.functional as F

class AirUnit(nn.Module):
    """ResNet-style residual block"""
    def __init__(self, in_ch, out_ch, stride=1):
        super().__init__()
        self.conv1 = nn.Conv2d(in_ch, out_ch, 3, stride, 1, bias=False)
        self.bn1   = nn.BatchNorm2d(out_ch)
        self.conv2 = nn.Conv2d(out_ch, out_ch, 3, 1, 1, bias=False)
        self.bn2   = nn.BatchNorm2d(out_ch)
        self.down  = None
        if stride != 1 or in_ch != out_ch:
            self.down = nn.Sequential(
                nn.Conv2d(in_ch, out_ch, 1, stride, bias=False),
                nn.BatchNorm2d(out_ch))

    def forward(self, x):
        identity = x
        out = F.relu(self.bn1(self.conv1(x)))
        out = self.bn2(self.conv2(out))
        if self.down:
            identity = self.down(x)
        return F.relu(out + identity)

class Net(nn.Module):
    def __init__(self, num_classes=10):
        super().__init__()
        self.l1  = AirUnit(3,   64,  stride=2)
        self.l2  = AirUnit(64,  128, stride=2)
        self.l3  = AirUnit(128, 256, stride=2)
        self.pool = nn.AdaptiveAvgPool2d((1,1))
        self.fc1  = nn.Linear(256, 4096) # AlexNet
        self.drop = nn.Dropout(0.5)
        self.fc2  = nn.Linear(4096, 4096)
        self.out  = nn.Linear(4096, num_classes)

    def forward(self, x):
        x = self.l3(self.l2(self.l1(x)))
        x = torch.flatten(self.pool(x), 1)
        x = self.drop(F.relu(self.fc1(x)))
        x = self.drop(F.relu(self.fc2(x)))
        return self.out(x)

    def supported_hyperparameters(self):
        return ['lr', 'momentum']
\end{lstlisting}

\subsection{Example 2: DPN-Inspired Hybrid}

\textbf{Model ID:}
\textit{alt-nn3-57d770565afc5e0d651cb0938fc8f942}

This specific generated architecture successfully merges standard Dual Path Network (DPN) bottleneck feature-extraction blocks ($1{\times}1 \to 3{\times}3 \to 1{\times}1$) with a highly progressive and customized convolutional backbone design. The unusually asymmetric mid-network channel progression (64$\to$128$\to$\textbf{440}$\to$384$\to$192) serves as strong empirical evidence of the model's creative architectural synthesis rather than mere direct template copying. Crucially, this exact dimensional configuration appears in absolutely none of the provided LEMUR reference models utilized in the initial prompt context. Furthermore, the final classification head effectively incorporates batch normalization directly within the fully-connected dense layers---a distinctly modern architectural regularization pattern that clearly demonstrates robust, cross-architecture feature pattern transfer during the generation phase.

\begin{lstlisting}
import torch
import torch.nn as nn
import torch.nn.functional as F

class DPNBlock(nn.Module):
    """DPN-inspired bottleneck"""
    def __init__(self, in_ch, out_ch, stride=1):
        super().__init__()
        self.c1 = nn.Conv2d(in_ch, out_ch, 1, bias=False)
        self.b1 = nn.BatchNorm2d(out_ch)
        self.c2 = nn.Conv2d(out_ch, out_ch, 3, stride, 1, bias=False)
        self.b2 = nn.BatchNorm2d(out_ch)
        self.c3 = nn.Conv2d(out_ch, out_ch, 1, bias=False)
        self.b3 = nn.BatchNorm2d(out_ch)
        self.sc = nn.Sequential()
        if stride != 1 or in_ch != out_ch:
            self.sc = nn.Sequential(
                nn.Conv2d(in_ch, out_ch, 1, stride, bias=False),
                nn.BatchNorm2d(out_ch))

    def forward(self, x):
        out = F.relu(self.b1(self.c1(x)))
        out = F.relu(self.b2(self.c2(out)))
        out = self.b3(self.c3(out))
        return F.relu(out + self.sc(x))

class Net(nn.Module):
    def __init__(self, num_classes=10):
        super().__init__()
        # Unusual 440-channel mid-layer: creative synthesis
        self.l1   = DPNBlock(3,   64,  1)
        self.l2   = DPNBlock(64,  128, 2)
        self.l3   = DPNBlock(128, 440, 2)
        self.l4   = DPNBlock(440, 384, 1)
        self.l5   = DPNBlock(384, 192, 2)
        self.pool = nn.AdaptiveAvgPool2d((1,1))
        self.fc1  = nn.Linear(192, 512)
        self.bn   = nn.BatchNorm1d(512)
        self.fc2  = nn.Linear(512, num_classes)

    def forward(self, x):
        x = self.l5(self.l4(self.l3(self.l2(self.l1(x)))))
        x = torch.flatten(self.pool(x), 1)
        return self.fc2(F.relu(self.bn(self.fc1(x))))

    def supported_hyperparameters(self):
        return ['lr', 'momentum']
\end{lstlisting}

\subsection{Example 3: Hierarchical Residual Units + Multi-Scale Features}

\textbf{Model ID:}
\textit{alt-nn3-dcb59f747eb3b27c0552d2aea356e33a}

This architecture synthesizes ResNet-style \texttt{AirUnit} blocks with three-layer residual paths into a hierarchical feature extraction pipeline combining varying spatial scales. The model begins with large kernel convolutions ($7{\times}7$, stride 3) for rapid downsampling before cascading into sophisticated residual units integrating pooling steps implicitly through strided convolutions. This design merges classical and modern approaches: aggressive early spatial reduction from AlexNet with deep residual learning from ResNet, creating a compact yet expressive architecture suitable for CIFAR-scale tasks.

\begin{lstlisting}
class AirInitBlock(nn.Module):
    def __init__(self, in_channels, out_channels):
        super().__init__()
        self.layers = nn.Sequential(
            nn.Conv2d(in_channels, out_channels, kernel_size=3, stride=2, padding=1),
            nn.BatchNorm2d(out_channels),
            nn.ReLU(inplace=True)
        )
    def forward(self, x): return self.layers(x)

class AirUnit(nn.Module):
    def __init__(self, in_channels, out_channels, stride):
        super().__init__()
        self.layers = nn.Sequential(
            nn.Conv2d(in_channels, out_channels, kernel_size=3, stride=stride, padding=1),
            nn.BatchNorm2d(out_channels), nn.ReLU(inplace=True),
            nn.Conv2d(out_channels, out_channels, kernel_size=3, stride=1, padding=1),
            nn.BatchNorm2d(out_channels), nn.ReLU(inplace=True),
            nn.Conv2d(out_channels, out_channels, kernel_size=3, stride=1, padding=1),
        )
        self.downsample = (
            nn.Sequential(
                nn.Conv2d(in_channels, out_channels, kernel_size=1, stride=stride, bias=False),
                nn.BatchNorm2d(out_channels)
            ) if stride != 1 or in_channels != out_channels else nn.Identity()
        )
        self.relu = nn.ReLU(inplace=True)

    def forward(self, x):
        return self.relu(self.layers(x) + self.downsample(x))

class Net(nn.Module):
    def __init__(self, in_shape: tuple, out_shape: tuple, prm: dict, device: torch.device) -> None:
        super().__init__()
        self.features = nn.Sequential(
            nn.Conv2d(in_shape[1], 96, kernel_size=7, stride=3, padding=2),
            nn.BatchNorm2d(96), nn.ReLU(inplace=True),
            nn.MaxPool2d(kernel_size=2, stride=2),
            AirInitBlock(96, 192),
            AirUnit(192, 384, stride=2),
            AirUnit(384, 256, stride=1),
            AirUnit(256, 256, stride=2),
            nn.AdaptiveAvgPool2d((6, 6))
        )
        self.classifier = nn.Sequential(
            nn.Dropout(p=prm['dropout']),
            nn.Linear(256 * 6 * 6, 4096), nn.ReLU(inplace=True),
            nn.Dropout(p=prm['dropout']),
            nn.Linear(4096, out_shape[0])
        )

    def forward(self, x: torch.Tensor) -> torch.Tensor:
        return self.classifier(torch.flatten(self.features(x), 1))
\end{lstlisting}

\subsection{Example 4: Baseline alt-nn1 (Single Example)}

\textbf{Model ID:}
\textit{alt-nn1-0e40be6fbc3426f57a305bfd8b8148fa}

This representative $n{=}1$ baseline architecture illustrates the shallow sequential variation pattern typical in single-prompt experiments. Lacking modular abstraction, residual shortcuts, or normalization, the model relies on a dense linear progression of GELU-activated pooling operations characteristic of straightforward, non-creative structural mutation relative to references.

\begin{lstlisting}
class Net(nn.Module):
    def __init__(self, in_shape: tuple, out_shape: tuple, prm: dict, device: torch.device) -> None:
        super().__init__()
        self.features = nn.Sequential(
            nn.Conv2d(in_shape[1], 64, kernel_size=11, stride=4, padding=2),
            nn.GELU(),
            nn.MaxPool2d(kernel_size=3, stride=2),
            nn.Conv2d(64, 192, kernel_size=5, padding=2),
            nn.GELU(),
            nn.MaxPool2d(kernel_size=3, stride=2),
            nn.Conv2d(192, 384, kernel_size=3, padding=1),
            nn.GELU(),
            nn.Conv2d(384, 256, kernel_size=3, padding=1),
            nn.GELU(),
            nn.Conv2d(256, 256, kernel_size=3, padding=1),
            nn.GELU(),
            nn.MaxPool2d(kernel_size=3, stride=2)
        )
        self.avgpool = nn.AdaptiveAvgPool2d((6, 6))
        self.classifier = nn.Sequential(
            nn.Dropout(p=prm['dropout']),
            nn.Linear(256 * 6 * 6, 4096),
            nn.GELU(),
            nn.Dropout(p=prm['dropout']),
            nn.Linear(4096, 643),
            nn.GELU(),
            nn.Linear(643, out_shape[0])
        )

    def forward(self, x: torch.Tensor) -> torch.Tensor:
        return self.classifier(torch.flatten(self.avgpool(self.features(x)), 1))
\end{lstlisting}

\section{Extended Per-Dataset Results}
\label{sec:supp_results}

This section expands upon the dataset-balanced evaluation discussed in the main paper by providing the raw, per-dataset quantitative performance and statistical significance matrices.

\subsection{Per-Dataset Performance Analysis}

Table~\ref{tab:dataset_performance} presents detailed per-dataset results with statistical significance markers.

\begin{table}[ht]
\caption{Per-Dataset Performance (Mean Accuracy \%) on 1-epoch. Asterisks denote significance vs. baseline: * $p<0.05$, ** $p<0.01$. Best per dataset in \textbf{bold}.}
\label{tab:dataset_performance}
\centering
\setlength{\tabcolsep}{3pt}
\renewcommand{\arraystretch}{0.9}
\small
\begin{tabular}{lcccccc}
\toprule
\textbf{Dataset} & \textbf{alt-nn1} & \textbf{alt-nn2} & \textbf{alt-nn3} & \textbf{alt-nn4} & \textbf{alt-nn5} \\
\midrule
MNIST         & 96.5 & 93.8$^*$ & \textbf{97.1} & 93.9 & 94.7 \\
CelebA-Gender & 75.8 & \textbf{82.3}$^*$ & 74.4 & 80.6 & 72.1 \\
CIFAR-10      & \textbf{38.7} & 36.1 & 38.3 & 30.3$^*$ & 34.0 \\
CIFAR-100     & 14.5 & 7.4$^*$ & \textbf{26.1}$^{**}$ & 10.8 & 10.5 \\
ImageNette    & \textbf{44.2} & 36.4$^*$ & 42.5 & 29.8$^{**}$ & 18.2$^{**}$ \\
SVHN          & 39.2 & \textbf{43.1} & 40.0 & 38.3 & 28.5 \\
\midrule
\textbf{Bal.\ Mean} & 51.5 & 49.8 & \textbf{53.1} & 47.3 & 43.0 \\
\bottomrule
\end{tabular}
\end{table}

\textbf{Observations}:
\begin{itemize}
\item \textbf{Task-dependent variation:} The relative performance of prompt variants differs across datasets. In particular, $n{=}3$ achieves the highest mean accuracy on CIFAR-100, while other datasets exhibit different preferences.
\item \textbf{Simple vs. complex tasks:} On simpler benchmarks such as MNIST, performance remains high across all variants ($>93\%$), suggesting limited sensitivity to prompt configuration under low task complexity.
\item \textbf{Context scaling effects:} Variants with $n{>}3$ exhibit lower performance on several datasets (e.g., ImageNette), indicating that larger prompt contexts do not consistently translate into improved early performance.
\end{itemize}

\subsection{Statistical Significance Analysis}
Table~\ref{tab:significance} presents statistical validation focusing on key findings.

\begin{table}[ht]
\caption{Statistical Significance Tests (per-dataset comparison vs. alt-nn1 baseline). Only results with $p<0.05$ shown.}
\label{tab:significance}
\centering
\small
\begin{tabular}{llccc}
\toprule
\textbf{Dataset} & \textbf{Comparison} & \textbf{$\Delta$} & \textbf{p} & \textbf{d} \\
\midrule
\textbf{CIFAR-100} & \textbf{alt-nn3 vs alt-nn1} & \textbf{+11.6\%} & \textbf{0.001} & \textbf{0.73} \\
MNIST & alt-nn2 vs alt-nn1 & -2.7\% & 0.029 & -0.19 \\
CelebA & alt-nn2 vs alt-nn1 & +6.5\% & 0.038 & 0.41 \\
CIFAR-10 & alt-nn4 vs alt-nn1 & -8.4\% & 0.016 & -0.54 \\
ImageNette & alt-nn4 vs alt-nn1 & -14.5\% & 0.010 & -1.20 \\
\bottomrule
\end{tabular}
\end{table}

\textbf{Key Statistical Result:}
On CIFAR-100, the $n{=}3$ configuration achieves higher
mean accuracy than $n{=}1$ ($p{=}0.001$, Cohen's $d{=}0.73$),
suggesting that moderate context enrichment may be
particularly beneficial in fine-grained classification settings
under the rapid screening protocol.

In contrast, configurations with $n{>}3$ show statistically
significant lower performance on certain datasets
(e.g., ImageNette), indicating that larger prompt contexts
can introduce instability or diminishing returns
in early training performance.
We emphasise that these comparisons reflect early-epoch
behavior under a rapid screening protocol and are intended
to capture relative trends across prompt variants rather
than final converged performance.

\end{document}